\documentclass[runningheads]{llncs}

\usepackage[utf8]{inputenc} 
\usepackage[T1]{fontenc}    
\usepackage{lmodern}
\usepackage{hyperref}       
\usepackage{url}            
\usepackage{booktabs}       
\usepackage{nicefrac}       
\usepackage{microtype}      

\usepackage{latexsym,amssymb,amstext,amsfonts,amsmath,setspace,mathtools,bm}
\usepackage{subcaption}
\usepackage{enumerate} 
\usepackage{paralist}
\usepackage{float}
\usepackage{array}
\usepackage{comment}

\usepackage{algorithm}
\usepackage{algpseudocode}
\usepackage{multicol}
\usepackage{lipsum}
\usepackage{xcolor}

\usepackage{tikz}
\usetikzlibrary{fit,positioning}
\usetikzlibrary{calc}

\definecolor{nice_blue}{RGB}{65, 105, 225}
\definecolor{nice_red}{RGB}{168, 34, 34}
\definecolor{IGCBlue}{HTML}{16197A}
\definecolor{dark_green}{RGB}{20, 110, 10}

\definecolor{graph_blue}{RGB}{144, 195, 212}
\definecolor{graph_purple}{RGB}{195, 144, 212}
\definecolor{graph_green}{RGB}{161, 212, 144}
\definecolor{graph_orred}{RGB}{212, 161, 144}

\usepackage{graphicx}

\begin{document}
\title{Does SAM dream of EIG? Characterizing Interactive Segmenter Performance using Expected Information Gain}
\titlerunning{Does SAM dream of Information?}
%

\author{Kuan-I Chung \and Daniel Moyer}

%
\authorrunning{K.Chung and D. Moyer}

%

\institute{Vanderbilt University, Nashville TN, USA\\
\email{\{kuan-i.chung, daniel.moyer\}@vanderbilt.edu}\\}


%
\maketitle              
\begin{abstract}
We introduce an assessment procedure for interactive segmentation models. Based on concepts from Bayesian Experimental Design, the procedure measures a model's understanding of point prompts and their correspondence with the desired segmentation mask. We show that Oracle Dice index measurements are insensitive or even misleading in measuring this property. We demonstrate the use of the proposed procedure on three interactive segmentation models and subsets of two large image segmentation datasets. 

\end{abstract}

\section{Introduction}
\label{sec:intro}

Interactive Segmentation methods allow humans to iteratively construct label maps through sequential prompting. Users annotate images by clicking points, drawing curves, or providing bounding boxes for the desired label annotation. Due to the ubiquity of segmentation tasks in medical imaging, recent work \cite{kirillov2023segany} on deep interactive segmenters have sparked great interest in the medical imaging community \cite{cheng2023sammed2d,deng2023segment,wong2023scribbleprompt,he2023computervision,hu2023sam,huang2024sam,mazurowski2023Sm,roy2023sammd,zhou2023sam,putz2023segment}; most are built upon the concepts and architecture introduced in the Segment Anything Model (SAM)\cite{kirillov2023segany}, originally intended for natural images. These studies measure performance through Oracle best-prompt or ad hoc heuristic prompt Dice index. 

For interactive segmentation methods, we contend that the oracle-prompt Dice index is at best half of the story. While high Dice under the optimal prompt is a necessary condition for good segmentation, good interaction requires an understanding of the information provided by a prompt about a segmentation. 
Specifically for point prompts, methods based on the SAM architecture provide their own implicit feedback about the quality of the potential next prompt, the expected information gain (EIG).
We believe the accuracy of this feedback speaks to how well a model parameterizes the interactions between proposed segmentations and user prompts for that given domain, as well as to the in-domain/out-of-domain shift of a model and its encapsulated priors.

In the present work we introduce a procedure to measure prompt/proposal understanding in interactive segmenter models. Our method makes use of Expected Information Gain, a quantity of interest taken from Bayesian Experimental Design. We frame the user-machine loop as iterative Bayesian updating of beliefs given observed data (pixel-wise segmentation probabilities, updated as the user provides inputs). We provide a procedure to measure performance under the information optimal decision, and we demonstrate this procedure using multiple interactive segmenters and datasets, showing that Expected Information Gain guidance discriminates well between models with varying point-prompt characteristics. Moreover, we show that when only using Oracle max Dice index, the methods are essentially indistinguishable. While previous work well captures model expressive capacity, we believe that EIG-based measurements characterize a model's understanding of point-prompt user interaction.

\subsection{Related Work}

Central to this work is the Segment Anything Model (SAM) \cite{kirillov2023segany} and similar and/or successor work in medical image segmentation \cite{cheng2023sammed2d,deng2023segment,wong2023scribbleprompt,he2023computervision,hu2023sam,huang2024sam,mazurowski2023Sm,roy2023sammd,zhou2023sam,putz2023segment}. The SAM model consists of an image encoder, a prompt encoder, and segmentation decoder. Images are fed into the model, and then users iterative provide prompts, resulting in proposal segmentations being output. This loops with users providing feedback on the proposal segmentation until convergence or failure.

A large subset of medical imaging studies assess SAM as an out-of-the-box segmentation model \cite{cheng2023sammed2d,deng2023segment,wong2023scribbleprompt,he2023computervision,hu2023sam,huang2024sam,mazurowski2023Sm,roy2023sammd,zhou2023sam}. Some directly employed human to interact with SAM~\cite{hu2023sam,huang2024sam}, while others simulate human behavior with heuristic procedures, such as focusing on boundaries, or correcting the wrongly segmented areas \cite{cheng2023sammed2d,he2023computervision,mazurowski2023Sm,putz2023segment,wong2023scribbleprompt}. A subset also test on random clicks \cite{deng2023segment,roy2023sammd,wong2023scribbleprompt} or use oracle procedures to find the  maximum possible Dice\cite{mazurowski2023Sm,putz2023segment}.

Clearly human interaction \cite{hu2023sam,huang2024sam} is the best and eventually the ``real'' metric of usefulness for an interactive segmenter. However, collecting these measurements is both expensive and prone to strong biases, such as differing levels of understanding of the interactive segmenter. Even if a skilled user is able to push an interactive segmenter to provide the correct mask, if they need to use idiosyncratic and uninituitive prompts to receive such a high accuracy we would still judge the interactive method to be a failure.

Our method is based on concepts from Bayesian Experimental Design \cite{foster2020vboed,rainforth2018nmc,rainforth2023modern}, which aims to adaptively choose ``designs'' or actions (here, the next prompt) that maximizes the information gained about target parameters (here, the segmentation masks).

\section{Methods}
\label{sec:methods}

\subsection{Information Gain and Optimal Design}
For an estimation objective $\theta$ and related observations $y$ at locations $d$, we would say that $(y,d)$ was a useful observation if it reduced the uncertainty about $\theta$. This intuition is operationalized as the measured change in entropy, which is equivalent to the Information Gain (IG):
\begin{equation}
    IG(y, d)=\underbrace{H[p(\theta)]}_{\text{prior entropy}}
- \underbrace{H[p(\theta|y;d)]}_{\text{posterior entropy}}\label{eq:ig}
\end{equation}
Here the ``posterior'' is the distribution of $\theta$ after observing the additional $(y,d)$ data point.
Unfortunately, IG requires us to observe $y$ at $d$ before we can measure how much information we gain from it. 

Instead, we measure the Expected IG (EIG) which is information we expect to gain without actually observing $(y,d)$, using our current beliefs about the distribution of $(y,d)$, $\theta$:
\begin{align}
EIG(d) =&\mathbb{E}_{p(y|d)}\big[H[p(\theta)]-H[p(\theta|y;d)]\big] \label{eq:theta-eig}
\end{align}
$EIG(d)$ can be manipulated to the following form \cite{foster2020vboed,rainforth2018nmc,rainforth2023modern}:
\begin{align}
EIG(d) 
=&\mathbb{E}_{p(\theta,y|d)}\Big[\log\frac{p(\theta,y|d)}{p(\theta)}\Big] \\
=&\mathbb{E}_{p(\theta,y|d)}\Big[\log p(y|\theta; d) - \log\big[\mathbb{E}_{p(\theta)}p(y|\theta;d)\big]\Big]\label{eq:y-eig}
\end{align}
We can then define the optimal design~$d^*$ as the one that maximizes the~$EIG(d)$.
\begin{equation}
    d^*=\arg\max_{d\in\mathcal{D}}EIG(d)  \label{eq:argmaxEIG}
\end{equation}
Importantly, optimality of $d$ is conditional on correct specification of beliefs, and appropriateness/closeness of priors. While the data generating process and the observational mechanisms may have intrinsic stochasticity, many $p(\theta|y;d)$ estimators such as deep neural networks carry with them strong structural priors and inductive biases. This makes them efficient in structured but high-dimensional domains such as images, but also means that they may make strongly sub-optimal choices with respect to practical outcomes (distance of mean $\theta$ or $p(\theta)$ to the ground-truth) even when selecting the max-EIG observation location $d^*$.

\subsection{Measurement of EIG in Interactive Segmenters}
\label{sec:fakeprior}

Interactive Segmenters such as SAM~\cite{kirillov2023segany} are purposefully built around iterative interactions: they receive data, produce beliefs (proposal segmentations), and receive additional data. This means that naturally fit into Bayesian Experimental Design settings, though users may not be taking the optimal action at each step.

We model each segmentation as the belief map $\theta$, which is the likelihood that pixel labels $y$ are from the positive class (chosen arbitrarily from the categorical segmentation labels). That is, $\theta_{ij} = p(y_{ij}=1)$. Each point prompt is a observation/location pair $(y,d)$, where $d$ is a pixel coordinate $(i,j)$. The (deep) segmentation model itself provides beliefs $\theta$ given $(y,d)$.





Using the above setup, we can produce a pixel-wise map of the EIG for a given $p(\theta)$ (recovered from the output of the interactive segmenter). We simplify our observational model by omitting any observer noise for $y$, so any observed pixel label is assumed to be correct. As integration over $p(\theta,y|d)$ and $p(\theta)$ is analytically intractable in general, we use a nested Monte Carlo scheme to estimate the necessary quantities. From Eq. \ref{eq:y-eig} we can then write the Monte Carlo approximation:
\begin{equation}
    EIG(d)\approx\frac{1}{N}\sum_{n=1}^N\Big[\log p(y_n|\theta_{n,0};d)-\log\big[\frac{1}{M}\sum_{m=1}^Mp(y_n|\theta_{n,m};d)\big]\Big], \label{eq:NMC}
\end{equation}
where~$\theta_{n,m}\sim p(\theta)$ for~$(m,n)\in\{0, ..., M\}\times\{1, ..., N\}$, ~$y_n\sim p(y|\theta=\theta_{n,0},d)$, and $N$ and $M$ are the sample sizes for outer and inner Monte Carlo loops, respectively. Rainforth et al.~\cite{rainforth2018nmc} shows that the left and right-hand sides of Eq. \ref{eq:NMC} are asymptotically equal as $N$ and $M$ grow large if $\sqrt{N} \propto M$.

We can simplify this further for categorical segmentation models, using the marginal Bernoulli likelihood of a given pixel label:
\begin{equation}
    p(y_n|\theta_{n,m},d)={\theta_{n,m}}^{y_n}(1-\theta_{n,m})^{1-y_n}. \label{eq:likelihood}
\end{equation}
The nested Monte Carlo then becomes:
\begin{equation}
\frac{1}{N}\sum_{n=1}^N
\bigg[\log\left({\theta_{n,0}}^{y_n}(1-\theta_{n,0})^{1-y_n}\right) -
\log\left(\frac{1}{M} \sum_{m=1}^M {\theta_{n,m}}^{y_n}(1-\theta_{n,m})^{1-y_n}\right)
 \bigg]. \label{eq:eig}
 \end{equation}
For more complex pixel labels the corresponding standard Multinomial/Dirichet distributions can be used here instead, though for simplicity in this paper we only experiment with binary segmentations. 



In order to estimate Eq. \ref{eq:eig} we need samples $\theta_{n,m}$; unfortunately, in general we receive only a point estimate of $p(\theta)$ from an interactive segmenter. However, the SAM architecture actually provides multiple of these estimates. We may then use the set of these prediction heads as a proxy $\theta$ distribution $q(\theta)= \text{Unif}(\{\theta_k\})$, where $k$ indexes the architecture defined heads. 

The procedure for sampling~$\theta_{n,m},y_n|d=(i,j)$ is then:
\begin{itemize}
    \item[1.] At the outer~$n$-th loop of the Monte Carlo, we sample $\theta_{n,0}$ from~$q(\theta)$ for outer Monte Carlo calculation.
    \item[2.] Sample~$p\sim Unif(0,1)$, then sample $y_n$:
              $$y_n=\begin{cases}
                1 \text{, if } p < (\theta_{n,0})_{i,j}\\
                0 \text{, else }
              \end{cases}.$$
              Note that this~$y_n$ is also used in inner Monte Carlo calculation.
    \item[3.] For the inner~$m$-th loop, sample~$\theta_{n,m}\sim q(\theta)$ for inner Monte Carlo calculation.
\end{itemize}

Importantly, these estimates put all spatial auto-correlation into the $\theta$ distribution, and assumes we will observe $y_{ij}$ one at a time in sequence. 
Thus, our $EIG(d)$ nested Monte Carlo estimate gives us the expected information gain at each pixel $(i,j)$ were it to be individually observed. Without loss of generality, we can then calculate all designs~$d\in \mathcal{D}$ simultaneously in each sampled~$\theta_{n,m}$, as they are separable.


%


\section{Experiments}
\label{sec:exp}

We experiment on three models (SAM, MedSAM, and SAM-Med2D), testing both their overall capacity (via Oracle Greedy Dice index) as well as their max-EIG performance. We examine performance on sub-categories of image segmentation tasks from two datasets (Microsoft COCO and SA-Med2D-20M), which provides variation between in-domain and out-of-domain for the three models.

\subsubsection{Models:}

\textbf{SAM:} We use original Segment Anything Model~\cite{kirillov2023segany} (SAM) with pretrained weights. Since the following medical SAMs are using ViT-b architecture, we use the same architecture here for comparison. \textbf{MedSAM}: ~\cite{ma2024medsam} was the first medical SAM model, which as exactly the same architecture to the original one's. It was entirely trained on medical images. However, this model only trained on bounding box prompts.\footnote{Our results are not meant to be an unfair indictment of the MedSAM training; we expect its EIG performance to be poor as they had no point prompt training.} \textbf{SAM-Med2D}~\cite{cheng2023sammed2d} has ViT-b architecture with additional adapter layer in image encoder. It was trained on SA-Med2D-20M datasets, including multiple public medical datasets. Besides bounding-box prompts, it was also trained with point prompts and mask prompts. In our experiments, we use SAM-Med2D with adapter.


\subsubsection{Datasets:}

The \textbf{Microsoft COCO} ~\cite{lin2015microsoft} dataset is a well-known large-scale dataset, collecting tons of natural images with object detection and segmentation task. In our experiments, we choose three categories: \textit{person}, \textit{car}, and \textit{keyboard} for inference.
\textbf{SA-Med2D-20M} ~\cite{ye2023samed2d20m} is a composite of multiple pre-processed public medical datasets. Each sample contains a 2D image and a 2D mask . Specifically, we choose three datasets for inference: \textit{FLARE}(abdomen CT scan)~\cite{ma2023flare22}, \textit{ACDC}(cardiac MRI)~\cite{bernard2018acdc}, and \textit{PALM}(fundus photo)~\cite{fu2019palm}.
\subsubsection{Prompt Simulation: }


As described in Section \ref{sec:fakeprior}, we use the ground truth label to simulate annotators providing noiseless pixel-wise observations of the desired label masks. To reduce the computational load, we use a $30\times30$ grid of possible annotation points for all experiments. We conduct two experiments, one querying pixels using the max-EIG pixel (``EIG-guided''), and the other searching over pixels for the prompt that best improves the Dice index (``Oracle''). We initialize both sequences of prompts with the center of the ground truth mask.

\textbf{EIG-guided: } Using Eq. \ref{eq:argmaxEIG}, we choose the optimal location $d^*=(i^*,j^*)$ as the next added point prompts and measure proposed mask accuracy. We then recalculate the EIG map for the next step, and loop.

\textbf{Oracle: } The Oracle sequence is constructed from the best next prompt. It clearly requires knowledge of the model output given an annotated trial point, and thus it is not a plausible real sequence (a user would have to try each grid point). However, these sequences measure the overall capacity of the encoder/decoder/prompt-encoder setup to express the correct mask. The Oracle step is locally the maximal Dice achievable, though we do not compute overall maximum Dice which would require searching over all configurations of prompts (which is prohibitively expensive computationally).

\subsection{Results}

\begin{figure}[!bth]
    \centering
    \includegraphics[width=1.0\linewidth]{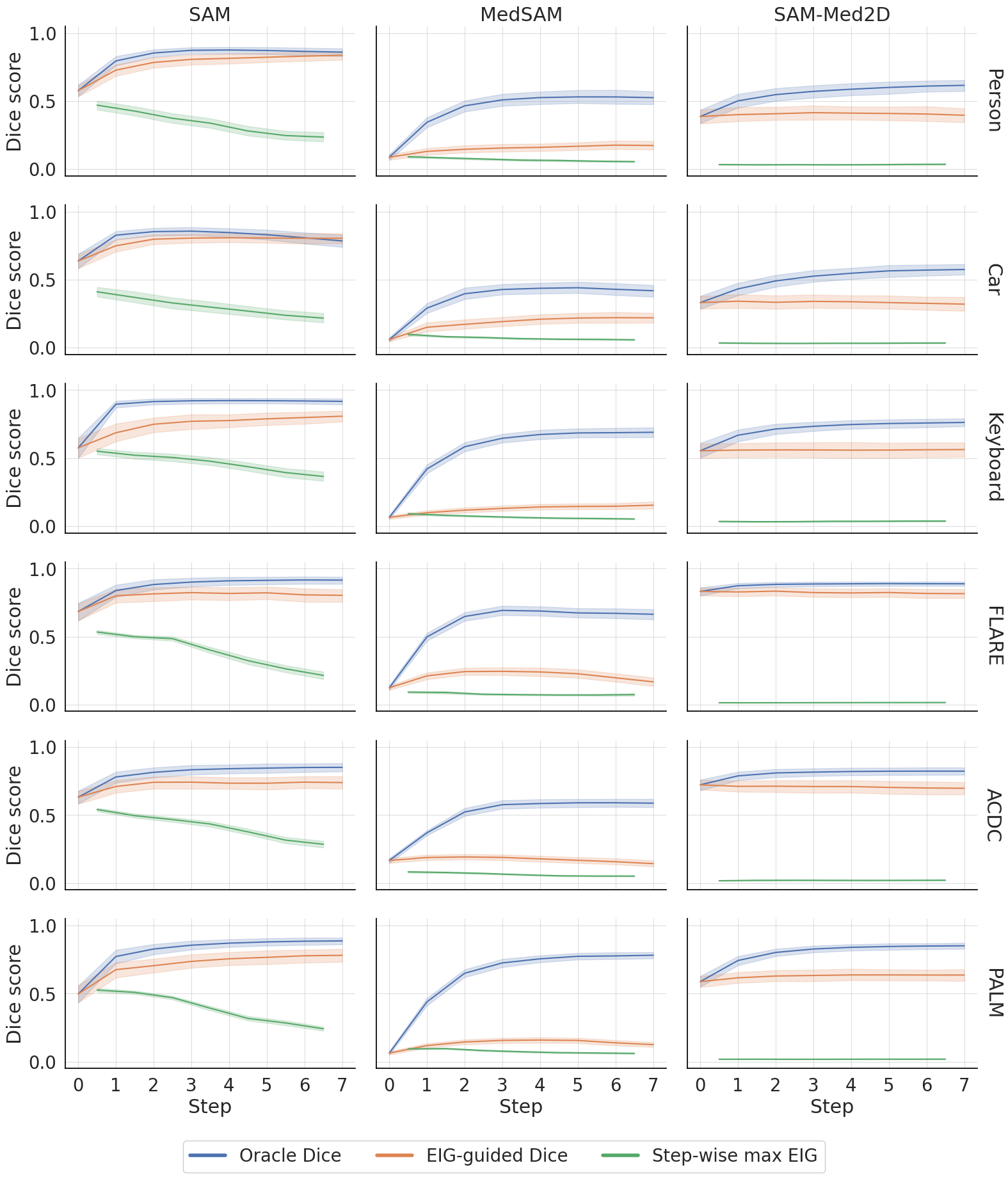}
    \caption{We display Oracle Dice index, EIG-Guided Dice index, and the step-wise max EIG amount (in Nats) as a function of steps (x-axis) for different SAM variants (columns) and categories of segmentation. Note that all quantities share the same y-axis scale; we plot the step-wise max EIG for context. }    
    \label{fig:BigFacetGrid}
\end{figure}

Figure \ref{fig:BigFacetGrid} shows the performance of different SAM's testing on different datasets. As can be seen in the plot, the original SAM method performs well on both natural images and medical images for both Oracle and EIG-guided measures, with a slight performance decrease on the ``out-of-domain'' medical imaging datasets. This reflects its relatively high capacity for both expressing different mask shapes, as well as understanding user prompting. Unsurprisingly, MedSAM has extremely weak EIG-Guided performance. This is expected as they \textbf{did not train their prompt encoder for point prompts}; we include the model not as unfair criticism to MedSAM, but to demonstrate that EIG-guidance is measuring point prompt understanding (or lack thereof). Finally, SAM-Med2D has very high Oracle segmentation performance, but surprisingly flat EIG-guided performance. This may reflect the models propensity to output more extreme statistical values.

\begin{figure}[t]
    \centering
    \includegraphics[width=1.0\linewidth]{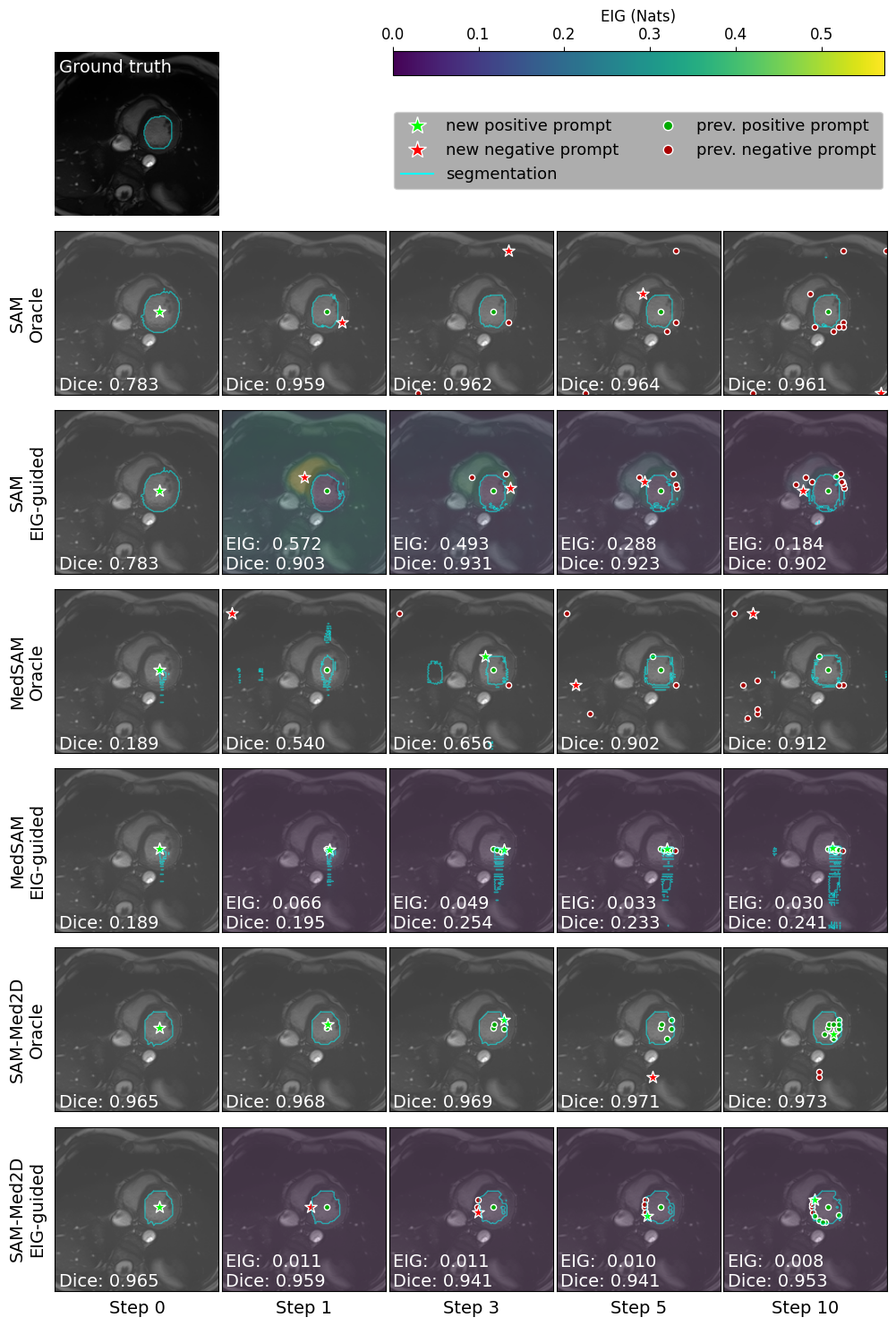}
    \caption{\textbf{Simulation}.
    The heat maps are the EIG before clicking the new prompts (stars). Hence, there's no EIG map for step 0. Essentially, the star signs are at highest EIG location. We also show the maximum EIG and the Dice score of prediction on the images.
    }
    \label{fig:ACDC_Compare}
\end{figure}

In Figure \ref{fig:ACDC_Compare} we visualize the point selection of the EIG-Guided and Oracle sequences for the different methods on an exemplar heart segmentation. The background heatmap on EIG-Guided images denotes the EIG map for the next step. The left most initial prompt is the same across all images, the center of the target segmentation. As expected, the MedSAM Oracle guidance increases Dice but is visually irrelevant to the target mask. Moreover, the input sequence is implausible as a user input. The SAM-Med2D model achieves very high accuracy after a single click, but subsequent clicks are very close spatially, and provide little additional information. For all three Oracle masks, the quality is relatively high (>0.9 Dice, and >0.95 in two cases); this demonstrates the utility of the EIG-guidance measure.


\section{Discussion and Conclusion}
\label{sec:disc}
This work utilizes expected information gain(EIG) as a implicit feedback from interactive segmenters. Results shows that it is be a meaningful metric for evaluating a models understanding of point prompts. We strongly believe that looking at only aggregate performance of Oracle best-prompts or random point prompts will not provide a comprehensive assessment of quality of an interactive segmentation model. Oracle measurements in particular can provide very high performance metrics while using completely implausible point sequences. We hypothesize this behavior is due to the extreme flexibility of the prompt encoder, where very unintuitive prompt points may lead to the locally optimal Dice gain. Instead, we feel that EIG and EIG-guided Dice index is a natural measurement of a model's understanding of user prompting, which is an essential part of assessing interactive segmentation.

\subsection{Acknowledgements}

Withheld for review.




\clearpage
\bibliographystyle{splncs04}
\bibliography{biblio.bib}

\end{document}